\documentclass{article}

\usepackage{microtype}
\usepackage{graphicx}
\usepackage{subfigure}
\usepackage{booktabs} % for professional tables

\usepackage{hyperref}

\usepackage[accepted]{icml2025}

\usepackage{amsmath}
\usepackage{amssymb}
\usepackage{mathtools}
\usepackage{amsthm}

\usepackage[capitalize,noabbrev]{cleveref}

\theoremstyle{plain}

\theoremstyle{definition}

\theoremstyle{remark}

\usepackage[textsize=tiny]{todonotes}

\icmltitlerunning{Sparse Autoencoders Trained on the Same Data Learn Different Features}

\begin{document}

\twocolumn[
\icmltitle{Sparse Autoencoders Trained on the Same Data Learn Different Features}

% It is OKAY to include author information, even for blind
% submissions: the style file will automatically remove it for you
% unless you've provided the [accepted] option to the icml2025
% package.

% List of affiliations: The first argument should be a (short)
% identifier you will use later to specify author affiliations
% Academic affiliations should list Department, University, City, Region, Country
% Industry affiliations should list Company, City, Region, Country

% You can specify symbols, otherwise they are numbered in order.
% Ideally, you should not use this facility. Affiliations will be numbered
% in order of appearance and this is the preferred way.
\icmlsetsymbol{equal}{*}

\begin{icmlauthorlist}
\icmlauthor{Gonçalo Paulo}{eai}
\icmlauthor{Nora Belrose}{eai}
\end{icmlauthorlist}

\icmlaffiliation{eai}{EleutherAI}

\icmlcorrespondingauthor{Gonçalo Paulo}{gonçalo@eleuther.ai}

% You may provide any keywords that you
% find helpful for describing your paper; these are used to populate
% the "keywords" metadata in the PDF but will not be shown in the document
\icmlkeywords{Machine Learning, ICML}

\vskip 0.3in
]

% this must go after the closing bracket ] following \twocolumn[ ...

% This command actually creates the footnote in the first column
% listing the affiliations and the copyright notice.
% The command takes one argument, which is text to display at the start of the footnote.
% The \icmlEqualContribution command is standard text for equal contribution.
% Remove it (just {}) if you do not need this facility.

\printAffiliationsAndNotice{}  % leave blank if no need to mention equal contribution
%\printAffiliationsAndNotice{\icmlEqualContribution} % otherwise use the standard text.

\begin{abstract}
Sparse autoencoders (SAEs) are a useful tool for uncovering human-interpretable features in the activations of large language models (LLMs). While some expect SAEs to find the true underlying features used by a model, our research shows that SAEs trained on the same model and data, differing only in the random seed used to initialize their weights, identify different sets of features. For example, in an SAE with 131K latents trained on a feedforward network in Llama 3 8B, only 30\% of the features were shared across different seeds. We observed this phenomenon across multiple layers of three different LLMs, two datasets, and several SAE architectures. While ReLU SAEs trained with the L1 sparsity loss showed greater stability across seeds, SAEs using the state-of-the-art TopK activation function were more seed-dependent, even when controlling for the level of sparsity. Our results suggest that the set of features uncovered by an SAE should be viewed as a pragmatically useful decomposition of activation space, rather than an exhaustive and universal list of features ``truly used'' by the model. % We provide evidence that this variability cannot be fully attributed to the recently discovered phenomenon of ``feature absorption,'' suggesting additional factors at play.
\end{abstract}

\section{Introduction}

Sparse autoencoders (SAEs) are an interpretability tool used to decompose neural network activations into human-understandable features \citep{cunningham2023sparse}. They address the problem of \emph{polysemanticity}, where individual neurons activate in semantically diverse contexts, defying any simple explanation \citep{arora2018linear,elhage2022toy}. SAEs consist of two parts: an encoder that transforms activation vectors into a sparse, higher-dimensional latent space, and a decoder that projects the latents back into the original space. Both parts are trained jointly to minimize reconstruction error. Recently, SAEs have been scaled tos state-of-the-art large language models, like GPT-4 \citep{gao2024scaling} and Claude 3 Sonnet \citep{templeton2024scaling}.

Many researchers hope SAEs can be used to ``identify and enumerate over all features in a model'' \citep{elhage2022toy}, which might allow us to check certain safety properties, such as that ``a model will never lie'' \citep{olah2023interpretability}. These hopes seem to presuppose that there is a unique, objective decomposition of a neural network into features, and that SAEs can uncover this decomposition \citep{smith2024strong}. In this paper, we test this presupposition by measuring the degree to which SAE features depend on the random seed used to initialize their weights.

It is somewhat nontrivial to compare features learned by different SAEs, since the latents have no inherent ordering. Given a trained SAE $\mathcal M$, we can generate a ``shuffled'' SAE $\mathcal M'$ by randomly permuting the rows of $\mathcal M$'s encoder matrix, and rearranging the columns of its decoder matrix using the inverse permutation.\footnote{This is generally true for any MLP with elementwise nonlinearities. Indeed, for ReLU networks there is also a continuous symmetry: the function represented by the model is unchanged when the pre-activation is scaled by $s$ and the post-activation is scaled by $s^{-1}$. The standard training recipe for SAEs eliminates this symmetry, however, by constraining decoder vectors to be unit norm \citep{bricken2023towards}.} $\mathcal M$ and $\mathcal M'$ represent the same function and contain \emph{the same features}, up to an irrelevant permutation symmetry, and yet their weights may look very different. Even if independently trained SAEs with different random initializations do learn the same features, we would expect their encoder and decoder vectors to be arranged in different orders.

We can get around this problem by computing a bijective \emph{matching} of each feature in the first SAE with a unique counterpart feature in second SAE. This matching should be ``optimal'' in the sense that it maximizes the average ``similarity'' between the matched features. This ensures that, in the case where one SAE has actually been generated by shuffling the features of another, we will conclude that the resulting SAEs do indeed have the same features. Luckily, the \href{https://en.wikipedia.org/wiki/Hungarian_algorithm}{Hungarian algorithm} is known to efficiently compute this optimal matching, and has been used to align independently trained networks before \citep{ainsworth2023git}.

\begin{figure}[t]
    \centering
    \includegraphics[width=1\linewidth]{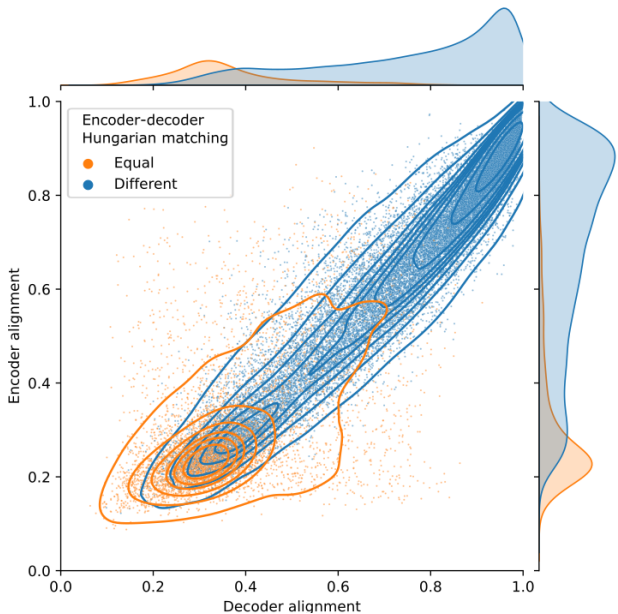}
    \caption{\textbf{Cosine similarities of features from SAE 1 with their counterparts in SAE 2.} Both SAEs have 32K latents, and are trained on the sixth MLP of Pythia 160M. Contour lines are regions of equal density according to kernel density estimation. We color each SAE 1 latent based on whether the Hungarian algorithm matches it to the same counterpart in SAE 2, or to a different one, when using the decoder and encoder directions. % This gives a similar result to just using a threshold on either the encoder or decoder alignment, and a similar result to using a greedy alignment algorithm that matches each SAE latent with the one that has the highest cosine similarity.
    }
    \label{fig:sae_alignment}
\end{figure}

Similarity between features can be measured in different ways. In this paper, we use two measures: the cosine similarity between the encoder vectors, and the cosine similarity between the decoder vectors. Empirically, we find that the distribution of cosine similarities between the matched features has two distinct modes: high-similarity shared features which have a close counterpart in the other model, and low-similarity ``orphan'' features which have been matched with a relatively unrelated feature (Figure~\ref{fig:sae_alignment}). Qualitatively, shared features have a strong tendency to occur in semantically similar contexts and share similar explanations, while orphan features are usually semantically unrelated.

This bimodal distribution allows us to analyze the \emph{fraction} of features that are shared between two SAEs. For the largest model we tested, Llama 3 8B \citep{dubey2024llama}, \textbf{only 30\% of features are shared} across both seeds. We find that smaller models, and smaller SAEs trained on the same model, tend to have higher fractions of shared features. We also apply the automated interpretability pipeline of \citet{paulo2024automatically} to compare the interpretability of shared and orphan features. We find that orphan features are often quite interpretable, so that an individual SAE training run is likely ``missing out'' on a number of interpretable features.

\begin{figure}[h]
    \centering
    \includegraphics[width=1\linewidth]{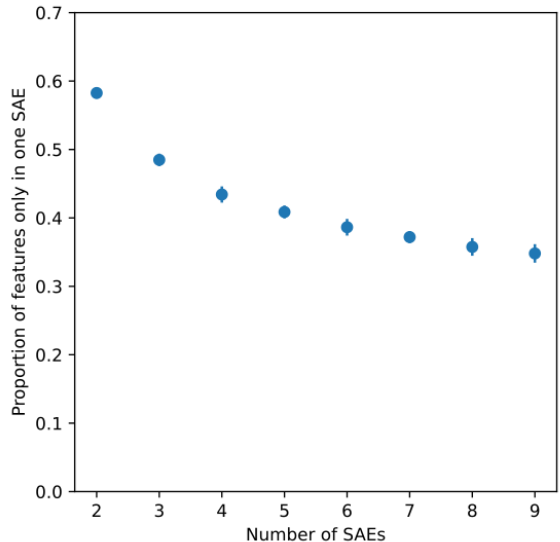}
    \caption{\textbf{Dependence of the number of latents found only in the base SAE on the number of seeds.} We consider a latent X in SAE A to be ``shared'' in SAE B if and only if X is matched to a latent Y in B with which it has cosine similarity greater than 0.7 according to both the encoder and decoder weights. To generate this plot we select a ``base'' SAE and compute its overlap with all the other seeds, then we average over all different base seeds. }
    \label{fig:multiple_seeds}
\end{figure}
\begin{figure*}[h]
    \centering
    \includegraphics[width=1\linewidth]{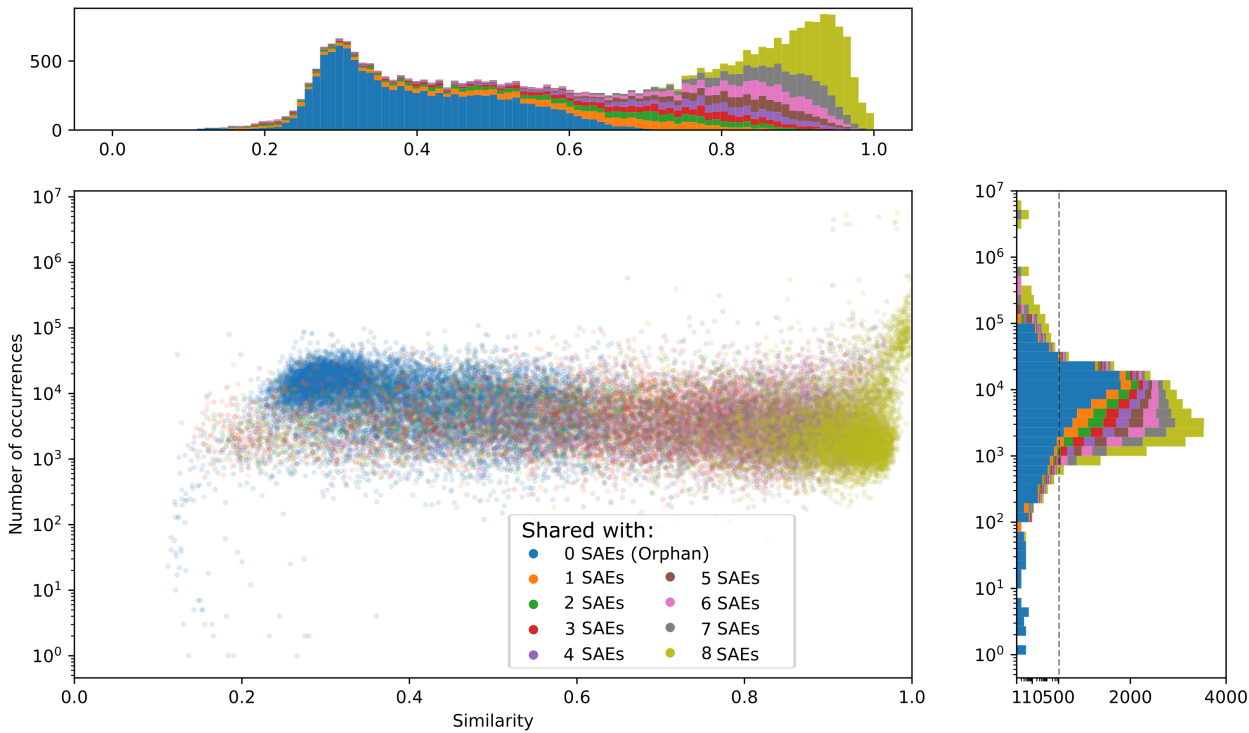}
    \caption{\textbf{Latent similarity vs. firing frequency.} We plot the cosine similarity between matched latents, vs. how often the latent fires in the base SAE. The similarity of each latent is averaged over all the matched latents of different seeds. The histograms in this figure are stacked, and the histogram of number of occurrences has a log-scale from 0 to 500, to highlight the few latents that rarely fire or that fire frequently, and a linear-scale from 500 to 4000. Latent occurrences were collected over 10M tokens of the Pile, the same dataset that the SAEs were trained on.
}
    \label{fig:overlap}
\end{figure*}
\section{Related Work}

Recent work has found that SAE features are not \emph{atomic}, in the sense that a ``meta SAE'' can decompose them into more specific features \citep{anonymous2024canonical}. Relatedly, a feature in a small SAE may be replaced by multiple, more specific features in a larger SAE. In some cases, a more general feature like \emph{starts with the letter L} appears alongside a specific feature like \emph{the token ``lion''}, which may prevent the general feature from being active in contexts where intuitively, both the general and the specific feature apply \citep{chanin2024absorption}. In light of these phenomena, some have questioned whether the ``flat'' design of standard SAEs can accommodate the hierarchical structure inherent to human concepts \citep{ayonrinde2024interpretability}. While sparse autoencoders presuppose that neural networks use linear representations, some research suggests that irreducibly nonlinear features also exist \citep{engels2024not}. If this is true, SAEs trained with different random initializations might converge to different ways of ``linearizing'' the nonlinear features in activation space.

Previous work had found that ReLU SAEs trained with an L1 sparsity penalty were stable under different seeds \citep{anonymous2024canonical,braun2024identifying}. By contrast, \citet{marks2024enhancing} found that TopK SAEs could be improved by training two different seeds and forcing them to be ``aligned,'' suggesting that they may not be sufficiently aligned by default. A recent benchmark of feature splitting showed a convergent result, where JumpReLU and TopK latents had a higher feature splitting rate than ReLU SAEs \citep{karvonen2024saebench}.

Concurrently with our work, \citet{balagansky2024mechanistic} use the Hungarian algorithm to align features from SAEs trained on adjacent layers of the Gemma 2 models \citep{gemmascope}. However, we learned in personal communication with the DeepMind interpretability team that the same random seed was used to initialize every SAE in the Gemmascope collection, so \citet{balagansky2024mechanistic}'s positive results are likely dependent on this hyperparameter choice.

\begin{table*}[h]
\begin{tabular}{|l|p{7.5cm}|p{7.5cm}|}
\hline
Alignment & Seed 1   & Seed 2   \\ \hline
0.10      & Abbreviated country name in United States Supreme Court case citations. $(0.865)$  & A single character or a small group of characters embedded within a larger word, often in a non-English language context (...). $(0.56)$  \\ \hline
0.27      & Definite articles and other words commonly used in formal and legal language, such as disclaimers, licensing terms, and court documents. $(0.91)$  & Punctuation marks or short words connecting or separating clauses, (...) and sometimes serving as conjunctions or prepositions. $(0.46)$ \\ \hline
0.44  & Abbreviated geographical or institutional references, usually in the context of legal citations. $(0.85)$   & Punctuation marks. $(0.49)$  \\ \hline
0.75 & Percent symbols marking numerical values representing proportions or rates. $(0.95)$ & A percentage symbol denoting the proportion of a quantity, (...) and usually in the form of a numerical value followed by the symbol. $(0.97)$ \\ \hline
0.97   & Adverbs that express frequency, such as 'often', 'sometimes', (...), used to indicate the occurrence or tendency of an event or action.  $(0.94)$ & Adverbs indicating frequency, such as 'often', 'frequently', (...), are used to describe the regularity or likelihood of an event or situation. $(0.99)$   \\ \hline
\end{tabular}
\caption{\textbf{Orphan latents can have high scoring explanations.} We selected explanations of pairs of latents shown in Fig. \ref{fig:scores_alignment}. Each explanation is shown alongside its detection score \citep{paulo2024automatically}, a number in $[0, 1]$ measuring explanation quality, in parentheses. We select latents from 5 bins of alignment by maximizing the score of both explanations if the cosine similarity between the latents is $>0.7$ and by maximizing the score of the explanation on seed 1 and minimizing the score on seed 2 of the cosine similarity is $<0.7$. Ellipsis added to some explanations for brevity. This choice was made to capture latents that had good explanations in seed 1 but were not matched in seed 2. The latent pairs are $(12314, 6024)$, $(21463, 3361)$, $(5888, 6649)$, $(14931, 5456)$ and $(1817, 66)$.  }
\label{tab:explanations}
\end{table*}
\section{Methods}
\label{sec:methods}

In order to measure the degree of alignment between independently trained SAEs, we use the Hungarian algorithm to efficiently compute the matching between their latents which maximizes the average cosine similarity between the matched encoder and decoder vectors. This average cosine similarity after matching is the overall alignment score. 

We begin by training two SAEs on the sixth MLP of Pythia 160M \citep{biderman2023pythia} with $2^{15}$ latents over the first 8B tokens of its own training corpus, the Pile \citep{gao2020pile}, using the \href{https://github.com/EleutherAI/sae}{\texttt{sae}} library \citep{belrose2024sae}. We use different random seeds for initialization, but both SAEs see exactly the same data in the same order. On this pair, we find that the distribution of matched cosine similarities has two modes: high-similarity ``shared'' features and low-similarity ``orphan'' features (Figure~\ref{fig:sae_alignment}). Overall, cosine similarities for encoder and decoder vectors are strongly correlated. We observe that in cases where the encoder and decoder matchings disagree (colored in orange), the cosine similarity is usually low for both matchings, whereas similarities are higher when the encoder and decoder matchings agree (colored in blue).

We use this to formally define a shared latent: \textbf{a latent is shared if it has the same counterpart latent in both the encoder and decoder matchings, \emph{and} in both of these matchings it has a cosine similarity of $0.7$ or greater}. Using this threshold, the fraction of shared latents is essentially unchanged if we were to use maximum cosine similarity (see below) in lieu of the Hungarian algorithm to label features as shared (Figure~\ref{fig:aligned_label}). By this definition, \textbf{only 42\% of latents are shared} across our two independently trained SAEs. If we consider that each latent of an SAE corresponds to a learned feature this result could mean that close to 60\% of learned features can be seed dependent.
% The threshold of $0.7$ also guarantees that nearly  has the consequence that Requiring that both the encoder and the decoder matchings lead to the same counterpart rules out 20\% of latents as candidates for being shared (Figure~\ref{fig:sae_alignment}, right panel orange line), but  $0.7$ or greater collapses all conditions to the same set of shared latent.

\paragraph{Maximum cosine similarity.} Prior work has measured the similarity of independently trained SAEs using the \emph{mean maximum cosine similarity} \citep[Figure 3b]{braun2024identifying}. Specifically, for each feature in the first SAE we find the maximum cosine similarity between itself and all features in the second SAE. The average of these maxima is the overall similarity score. This metric is simple, but it has the downside that it does not yield a bijective matching: many features in the first SAE may be mapped to one feature in the second SAE. We compare the matched cosine similarity produced by the Hungarian algorithm to the maximum cosine similarity for each latent in Figure~\ref{fig:matched_max_cosine_scatter}, observing that while for some latents the max cosine similarity is higher than the matched cosine similarity, the vast majority have the same value for both metrics, suggesting that the Hungarian algorithm has chosen to match most latents with their nearest neighbors. While we think the Hungarian matching approach is more principled and use it in the rest of this paper, we do find that empirically the difference between these two approaches is small.

\begin{figure*}[h]
    \centering
    \includegraphics[width=1\linewidth]{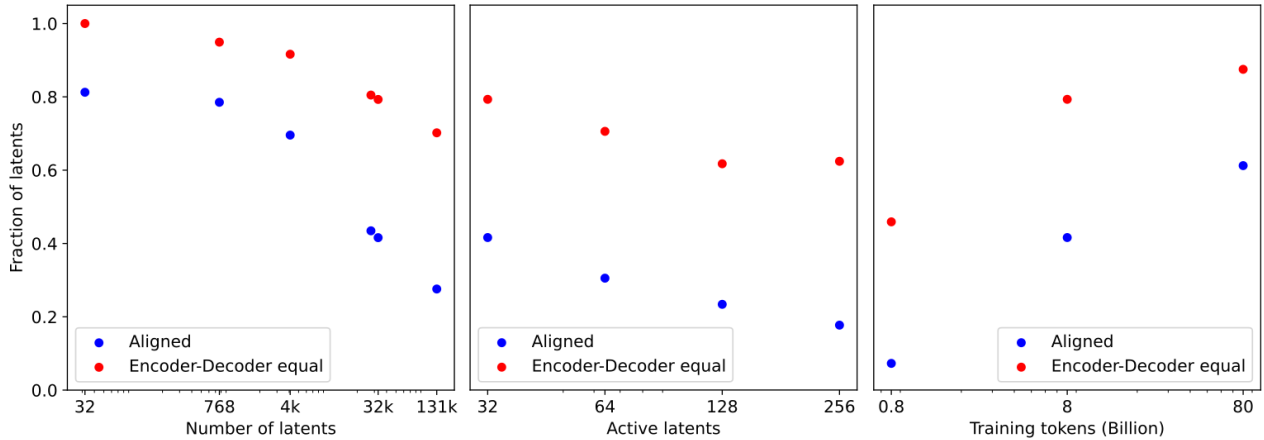}
    \caption{\textbf{Dependence of overlap of a Pythia-160M SAE on size, number of active latents and training time}. On the left we see that the fraction of aligned latents decreases with the increase of the number of latents. Middle shows that increasing the number of active latents, by increasing the value of $k$ for the TopK activation function, also decreases the overlap. On the right, training time increases the alignment of different SAE seeds.
    Unless otherwise indicated, each SAE has $2^{15}$ latents and was trained on the output of the sixth layer MLP of Pythia 160M, on the first 8B tokens of its training corpus, the Pile.
}
    \label{fig:hyperparameters}
\end{figure*}

\begin{figure*}[h!]
    \centering
    \includegraphics[width=1\linewidth]{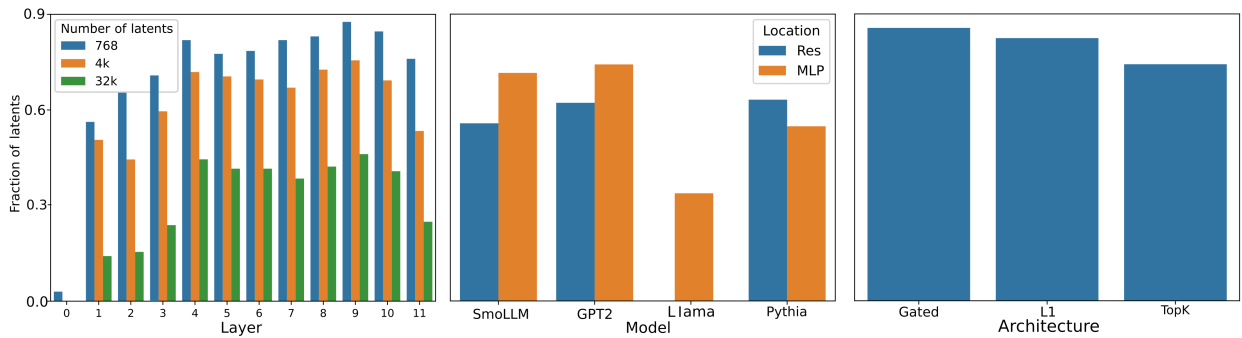}
    \caption{\textbf{Dependence of overlap on SAE hyperparameters.} On the right we see the how the fraction of shared latents for a Pythia-160M SAE depends on the layer and on the number of latents. In the middle we compare SAEs with the same expansion factor, 36, trained on different models and positions. On the right we compare SAEs trained on GPT2 using different activation functions and architectures. 
}
    \label{fig:hyperparameters_other}
\end{figure*}

\section{Asymptotic Trend}
 
We now consider seven more SAEs with the same data order, but with seeds different from the first two, yielding a total of nine independently trained SAEs. We first run the Hungarian algorithm $\binom{9}{2} = 36$ times, one for each pair of SAEs. Then tor each integer $k$ on the x-axis of Figure~\ref{fig:multiple_seeds}, we iterate over all $\binom{9}{k}$ combinations of SAEs of size $k$, and for each combination, we then run the following experiment $k$ different times, each time using a different SAE as the ``base SAE.'' We use each of the $k - 1$ matchings of the base SAE with a different SAE within this combination to compute a binary mask classifying each latent as shared or orphan, using the definition from Section~\ref{sec:methods}. We say that a latent is ``only in the base SAE'' if it is an orphan according to all $k - 1$ of these binary masks. Then, with respect to a given base SAE, we compute the proportion of all latents that are only in the base. Finally, we average the proportions generated by running this experiment $k \times \binom{9}{k}$ times, one for each combination of $k$ SAEs and each possible base SAE in each combination. The results of these experiments are plotted in Figure~\ref{fig:multiple_seeds}. When $k = 9$, we find that number of latents found in only one SAE is reduced to about 35\% (Figure~\ref{fig:multiple_seeds}, left panel).% We also observe a non-monotonic dependence of the number of SAEs that ``share'' any given latent - there is a minimum fraction of shared latents on exactly 5 seeds, this number increasing as we decrease or increase the number of seeds the latent is shared across.

The number of latents found in only one SAE decreases slowly as the number of seeds increases (Figure~\ref{fig:multiple_seeds}, right panel). Our results seems to indicate that when training a small number of SAEs a certain number of ``orphan'' latents are always present -- we found that a power law with an offset term fits the data significantly better than one without the offset.

To generate Figure~\ref{fig:overlap}, we fix Seed 1 as the base SAE, and color latents based on the number of matchings (out of the $9 - 1 = 8$ matchings involving Seed 1) in which they are classified as shared rather than orphan. We find that the latents that most frequently fire in the first SAE are the ones that are shared with all eight SAEs, and that the ones that most infrequently fire are the ones that are not shared with any other SAE, see Figure \ref{fig:overlap}. Interestingly, a significant number of orphan latents have a higher firing rate on average than latents that shared with all SAEs. In fact, as the average alignment between latents increases, the firing frequency seems to decrease.

\begin{figure*}[h!]
    \centering
    \includegraphics[width=1\linewidth]{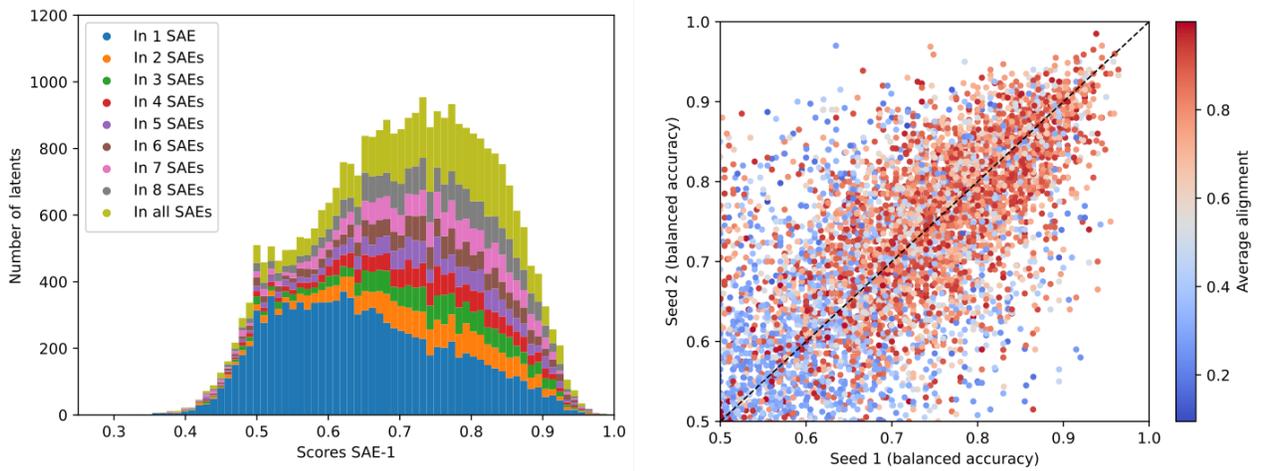}
    \caption{ \textbf{Interpretability of orphan features.} Distribution of scores of different latent explanation conditioned on the number of SAEs that latent can be found on. On the right we compare the scores of 5k explanations of matched latent of different SAE seeds. We see that most of the latents that have low alignment either have a low score or have a higher score in one of the SAEs than the other.}
    \label{fig:scores_alignment}
\end{figure*}

\section{Are ``Orphan'' Latents Interpretable?}

Intuitively, ``shared'' latents should have similar explanations to their counterparts and fire in semantically similar contexts, while ``orphan'' latents could be semantically different from their nearest counterparts. In this section, we generate explanations for all latents of two seeds of a $2^{15}$ latent SAE and score them using detection scoring \citep{paulo2024automatically}, evaluating the explanation over 100 active sequences and 100 non-active sequences. The average score of the explanations of the 32K SAEs is 0.72, with only 25\% of explanations having a score lower than 0.62, and only 25\% having a score better than 0.8.

Plotting the distributions of scores conditioned on the number of SAEs that ``shared'' that latent reveals that latents ``shared'' across a higher number of SAE seed have on average higher scores, meaning that these latents are more easily interpreted. In spite of this, a significant fraction of latents found only on one SAE, have high scoring latents. Plotting the scores of the latents of the two seeds mentioned above, we find that the most of the latents that have low similarity have either a low or an average score, see Figure \ref{fig:scores_alignment} left. Some latents have a average cosine alignment $<0.7$ and high scores, reinforcing the observation that some interpretable latents can be missing from any given seed, see Table \ref{tab:explanations} for some examples.

\section{Ablations}

We performed several ablation studies to investigate how our results depend on the hyperparameters used to train the SAE, including the number of active latents $k$, the total number of latents, the number of tokens used for training, and the SAE architecture (TopK, Gated, or ReLU).

We find that increasing the number of SAE latents, all else being equal, decreases the overlap between different seeds, see Figure \ref{fig:hyperparameters}. Increasing the number of active latents, by increasing the $k$ for TopK SAEs, also decreases the overlap, while the training time increases the overlap between latents. These results seem to indicate that what the seed dependence is not mainly due to feature absorption, as absorption increases when sparsity is decreased \citep{karvonen2024saebench}, and the model is trained for longer, while it does increase when the number of latents increases. We have found no evidence of feature absorption on the MLP SAEs we trained, but that may be due to the fact that the current metric is not tuned to find absorption on MLP SAEs, as it was mostly used on residual stream ones.  

The overlap between different seeds remains almost constant across the middle layers of the model, being lower for the earlier layers and the last layer (Figure~\ref{fig:hyperparameters_other}).  On SmolLM and GPT2 the MLP latents have more overlap between seeds than the residual stream ones, but the same is not true for Pythia. Previous work had found  that a large number of latents ($> 90\%$) where shared between GPT2 seeds \citep{anonymous2024canonical,braun2024identifying}, although those numbers where measured for SAEs with smaller numbers of latents than ours, and using a different architecture (ReLU instead of TopK). Indeed we find that standard and Gated SAEs \citep{rajamanoharan2024improving} trained with L1 loss have a larger overlap between latents. The overlap is much lower for the Llama 8B SAEs, which has more latents but the same expansion factor. 

In Figure~\ref{fig:matched_max_cosine} we compare the matched cosine similarity produced by the Hungarian algorithm to the maximum cosine similarity, showing that these are strongly correlated. This shows that our results are not strongly dependent on the choice of method used to compare features from independently trained SAEs.

\section{Conclusion}

Our results are further evidence for the idea that SAEs do not uncover a ``universal'' set of features. Different random initializations can lead to different sets of features being found, and SAEs seem to diverge, rather than converge, with increasing scale. We think feature discovery is best viewed as a compositional problem, wherein we look for useful ways of cutting up the activation space into categories, and these categories can themselves be cut up into further categories, hierarchically.

Mathematically, the lack of universality we observe here is due to the nonconvexity of the SAE loss function, which gives rise to many local optima. One might have expected a priori, however, that different local optima would have more feature overlap than we found in this study.

\section{Code Availability.}

Code to perform the Hungarian aligment of SAE seeds can be found
\href{https://anonymous.4open.science/r/sae_overlap-51F1}{here}.
%Code to perform the Hungarian alignment of SAE seeds as well as all the analysis done in this work can be found \href{https://github.com/EleutherAI/sae_overlap}{here}, and all the trained SAE checkpoints can be found \href{https://huggingface.co/EleutherAI/sae_overlap/tree/main}{here}.

%\section*{Acknowledgements}
% Gonçalo Paulo had the idea of investigating the overlap in features learned by SAEs trained with different random seeds, trained the SAEs, and ran all the experiments. Gonçalo wrote the first draft, and Nora Belrose extensively edited the draft into a final version.
 
% Gonçalo and Nora are funded by a \href{https://www.openphilanthropy.org/grants/eleuther-ai-interpretability-research/}{grant} from Open Philanthropy. We thank Coreweave for computing resources.

\section*{Impact Statement}

This paper presents work whose goal is to advance the field of Mechanistic Interpretability. There are many potential societal consequences 
of our work, none which we feel must be specifically highlighted here.

\bibliography{bibliography}

\begin{thebibliography}{24}
\providecommand{\natexlab}[1]{#1}
\providecommand{\url}[1]{\texttt{#1}}
\expandafter\ifx\csname urlstyle\endcsname\relax
  \providecommand{\doi}[1]{doi: #1}\else
  \providecommand{\doi}{doi: \begingroup \urlstyle{rm}\Url}\fi

\bibitem[Ainsworth et~al.(2023)Ainsworth, Hayase, and Srinivasa]{ainsworth2023git}
Ainsworth, S., Hayase, J., and Srinivasa, S.
\newblock Git re-basin: Merging models modulo permutation symmetries.
\newblock In \emph{The Eleventh International Conference on Learning Representations}, 2023.
\newblock URL \url{https://openreview.net/forum?id=CQsmMYmlP5T}.

\bibitem[Anonymous(2024)]{anonymous2024canonical}
Anonymous.
\newblock Sparse autoencoders do not find canonical units of analysis.
\newblock In \emph{Submitted to The Thirteenth International Conference on Learning Representations}, 2024.
\newblock URL \url{https://openreview.net/forum?id=9ca9eHNrdH}.
\newblock under review.

\bibitem[Arora et~al.(2018)Arora, Li, Liang, Ma, and Risteski]{arora2018linear}
Arora, S., Li, Y., Liang, Y., Ma, T., and Risteski, A.
\newblock Linear algebraic structure of word senses, with applications to polysemy.
\newblock \emph{Transactions of the Association for Computational Linguistics}, 6:\penalty0 483--495, 2018.

\bibitem[Ayonrinde et~al.(2024)Ayonrinde, Pearce, and Sharkey]{ayonrinde2024interpretability}
Ayonrinde, K., Pearce, M.~T., and Sharkey, L.
\newblock Interpretability as compression: Reconsidering sae explanations of neural activations with mdl-saes.
\newblock \emph{arXiv preprint arXiv:2410.11179}, 2024.

\bibitem[Balagansky et~al.(2024)Balagansky, Maksimov, and Gavrilov]{balagansky2024mechanistic}
Balagansky, N., Maksimov, I., and Gavrilov, D.
\newblock Mechanistic permutability: Match features across layers.
\newblock \emph{arXiv preprint arXiv:2410.07656}, 2024.

\bibitem[Belrose(2024)]{belrose2024sae}
Belrose, N.
\newblock Sae repository.
\newblock GitHub repository, 2024.
\newblock URL \url{https://github.com/EleutherAI/sae}.

\bibitem[Biderman et~al.(2023)Biderman, Schoelkopf, Anthony, Bradley, O’Brien, Hallahan, Khan, Purohit, Prashanth, Raff, et~al.]{biderman2023pythia}
Biderman, S., Schoelkopf, H., Anthony, Q.~G., Bradley, H., O’Brien, K., Hallahan, E., Khan, M.~A., Purohit, S., Prashanth, U.~S., Raff, E., et~al.
\newblock Pythia: A suite for analyzing large language models across training and scaling.
\newblock In \emph{International Conference on Machine Learning}, pp.\  2397--2430. PMLR, 2023.

\bibitem[Braun et~al.(2024)Braun, Taylor, Goldowsky-Dill, and Sharkey]{braun2024identifying}
Braun, D., Taylor, J., Goldowsky-Dill, N., and Sharkey, L.
\newblock Identifying functionally important features with end-to-end sparse dictionary learning, 2024.
\newblock URL \url{https://arxiv.org/abs/2405.12241}.

\bibitem[Bricken et~al.(2023)Bricken, Templeton, Batson, Chen, Jermyn, Conerly, Turner, Anil, Denison, Askell, Lasenby, Wu, Kravec, Schiefer, Maxwell, Joseph, Tamkin, Nguyen, McLean, Burke, Hume, Carter, Henighan, and Olah]{bricken2023towards}
Bricken, T., Templeton, A., Batson, J., Chen, B., Jermyn, A., Conerly, T., Turner, N.~L., Anil, C., Denison, C., Askell, A., Lasenby, R., Wu, Y., Kravec, S., Schiefer, N., Maxwell, T., Joseph, N., Tamkin, A., Nguyen, K., McLean, B., Burke, J.~E., Hume, T., Carter, S., Henighan, T., and Olah, C.
\newblock Towards monosemanticity: Decomposing language models with dictionary learning.
\newblock \emph{Transformer Circuits Thread}, 2023.
\newblock URL \url{https://transformer-circuits.pub/2023/monosemantic-features}.
\newblock Published October 4, 2023.

\bibitem[Chanin et~al.(2024)Chanin, Wilken-Smith, Dulka, Bhatnagar, and Bloom]{chanin2024absorption}
Chanin, D., Wilken-Smith, J., Dulka, T., Bhatnagar, H., and Bloom, J.
\newblock A is for absorption: Studying feature splitting and absorption in sparse autoencoders.
\newblock \emph{arXiv preprint arXiv:2409.14507}, 2024.

\bibitem[Cunningham et~al.(2023)Cunningham, Ewart, Riggs, Huben, and Sharkey]{cunningham2023sparse}
Cunningham, H., Ewart, A., Riggs, L., Huben, R., and Sharkey, L.
\newblock Sparse autoencoders find highly interpretable features in language models.
\newblock \emph{arXiv preprint arXiv:2309.08600}, 2023.

\bibitem[Dubey et~al.(2024)Dubey, Jauhri, Pandey, Kadian, Al-Dahle, Letman, Mathur, Schelten, Yang, Fan, et~al.]{dubey2024llama}
Dubey, A., Jauhri, A., Pandey, A., Kadian, A., Al-Dahle, A., Letman, A., Mathur, A., Schelten, A., Yang, A., Fan, A., et~al.
\newblock The llama 3 herd of models.
\newblock \emph{arXiv preprint arXiv:2407.21783}, 2024.

\bibitem[Elhage et~al.(2022)Elhage, Hume, Olsson, Schiefer, Henighan, Kravec, Hatfield-Dodds, Lasenby, Drain, Chen, et~al.]{elhage2022toy}
Elhage, N., Hume, T., Olsson, C., Schiefer, N., Henighan, T., Kravec, S., Hatfield-Dodds, Z., Lasenby, R., Drain, D., Chen, C., et~al.
\newblock Toy models of superposition.
\newblock \emph{arXiv preprint arXiv:2209.10652}, 2022.

\bibitem[Engels et~al.(2024)Engels, Michaud, Liao, Gurnee, and Tegmark]{engels2024not}
Engels, J., Michaud, E.~J., Liao, I., Gurnee, W., and Tegmark, M.
\newblock Not all language model features are linear.
\newblock \emph{arXiv preprint arXiv:2405.14860}, 2024.

\bibitem[Gao et~al.(2020)Gao, Biderman, Black, Golding, Hoppe, Foster, Phang, He, Thite, Nabeshima, et~al.]{gao2020pile}
Gao, L., Biderman, S., Black, S., Golding, L., Hoppe, T., Foster, C., Phang, J., He, H., Thite, A., Nabeshima, N., et~al.
\newblock The pile: An 800gb dataset of diverse text for language modeling.
\newblock \emph{arXiv preprint arXiv:2101.00027}, 2020.

\bibitem[Gao et~al.(2024)Gao, la~Tour, Tillman, Goh, Troll, Radford, Sutskever, Leike, and Wu]{gao2024scaling}
Gao, L., la~Tour, T.~D., Tillman, H., Goh, G., Troll, R., Radford, A., Sutskever, I., Leike, J., and Wu, J.
\newblock Scaling and evaluating sparse autoencoders.
\newblock \emph{arXiv preprint arXiv:2406.04093}, 2024.

\bibitem[Karvonen et~al.(2024)Karvonen, Rager, Lin, Tigges, Bloom, Chanin, Lau, Farrell, Conmy, McDougall, Ayonrinde, Wearden, Marks, and Nanda]{karvonen2024saebench}
Karvonen, A., Rager, C., Lin, J., Tigges, C., Bloom, J., Chanin, D., Lau, Y.-T., Farrell, E., Conmy, A., McDougall, C., Ayonrinde, K., Wearden, M., Marks, S., and Nanda, N.
\newblock Saebench: A comprehensive benchmark for sparse autoencoders, 2024.
\newblock URL \url{https://www.neuronpedia.org/sae-bench/info}.
\newblock Accessed: 2025-01-17.

\bibitem[Lieberum et~al.(2024)Lieberum, Rajamanoharan, Conmy, Smith, Sonnerat, Varma, Kramár, Dragan, Shah, and Nanda]{gemmascope}
Lieberum, T., Rajamanoharan, S., Conmy, A., Smith, L., Sonnerat, N., Varma, V., Kramár, J., Dragan, A., Shah, R., and Nanda, N.
\newblock Gemma scope: Open sparse autoencoders everywhere all at once on gemma 2, 2024.
\newblock URL \url{https://arxiv.org/abs/2408.05147}.

\bibitem[Marks et~al.(2024)Marks, Paren, Krueger, and Barez]{marks2024enhancing}
Marks, L., Paren, A., Krueger, D., and Barez, F.
\newblock Enhancing neural network interpretability with feature-aligned sparse autoencoders.
\newblock \emph{arXiv preprint arXiv:2411.01220}, 2024.

\bibitem[Olah(2023)]{olah2023interpretability}
Olah, C.
\newblock Interpretability dreams.
\newblock \emph{Transformer Circuits Thread}, 2023.
\newblock URL \url{https://transformer-circuits.pub/2023/interpretability-dreams/index.html}.
\newblock Published May 24, 2023.

\bibitem[Paulo et~al.(2024)Paulo, Mallen, Juang, and Belrose]{paulo2024automatically}
Paulo, G., Mallen, A., Juang, C., and Belrose, N.
\newblock Automatically interpreting millions of features in large language models.
\newblock \emph{arXiv preprint arXiv:2410.13928}, 2024.

\bibitem[Rajamanoharan et~al.(2024)Rajamanoharan, Conmy, Smith, Lieberum, Varma, Kram{\'a}r, Shah, and Nanda]{rajamanoharan2024improving}
Rajamanoharan, S., Conmy, A., Smith, L., Lieberum, T., Varma, V., Kram{\'a}r, J., Shah, R., and Nanda, N.
\newblock Improving dictionary learning with gated sparse autoencoders.
\newblock \emph{arXiv preprint arXiv:2404.16014}, 2024.

\bibitem[Smith(2024)]{smith2024strong}
Smith, L.
\newblock The strong feature hypothesis could be wrong, 2024.
\newblock URL \url{https://www.lesswrong.com/posts/tojtPCCRpKLSHBdpn/the-strong-feature-hypothesis-could-be-wrong}.
\newblock Accessed: 2025-01-17.

\bibitem[Templeton et~al.(2024)Templeton, Conerly, Marcus, Lindsey, Bricken, Chen, Pearce, Citro, Ameisen, Jones, Cunningham, Turner, McDougall, MacDiarmid, Freeman, Sumers, Rees, Batson, Jermyn, Carter, Olah, and Henighan]{templeton2024scaling}
Templeton, A., Conerly, T., Marcus, J., Lindsey, J., Bricken, T., Chen, B., Pearce, A., Citro, C., Ameisen, E., Jones, A., Cunningham, H., Turner, N.~L., McDougall, C., MacDiarmid, M., Freeman, C.~D., Sumers, T.~R., Rees, E., Batson, J., Jermyn, A., Carter, S., Olah, C., and Henighan, T.
\newblock Scaling monosemanticity: Extracting interpretable features from claude 3 sonnet.
\newblock \emph{Transformer Circuits Thread}, 2024.
\newblock URL \url{https://transformer-circuits.pub/2024/scaling-monosemanticity/index.html}.

\end{thebibliography}
\bibliographystyle{icml2025}

%%%%%%%%%%%%%%%%%%%%%%%%%%%%%%%%%%%%%%%%%%%%%%%%%%%%%%%%%%%%%%%%%%%%%%%%%%%%%%%
%%%%%%%%%%%%%%%%%%%%%%%%%%%%%%%%%%%%%%%%%%%%%%%%%%%%%%%%%%%%%%%%%%%%%%%%%%%%%%%
% APPENDIX
%%%%%%%%%%%%%%%%%%%%%%%%%%%%%%%%%%%%%%%%%%%%%%%%%%%%%%%%%%%%%%%%%%%%%%%%%%%%%%%
%%%%%%%%%%%%%%%%%%%%%%%%%%%%%%%%%%%%%%%%%%%%%%%%%%%%%%%%%%%%%%%%%%%%%%%%%%%%%%%
\newpage
\appendix
\onecolumn

\renewcommand{\thefigure}{A\arabic{figure}}

\setcounter{figure}{0}
\begin{figure}
    \centering
    \includegraphics[width=0.6\linewidth]{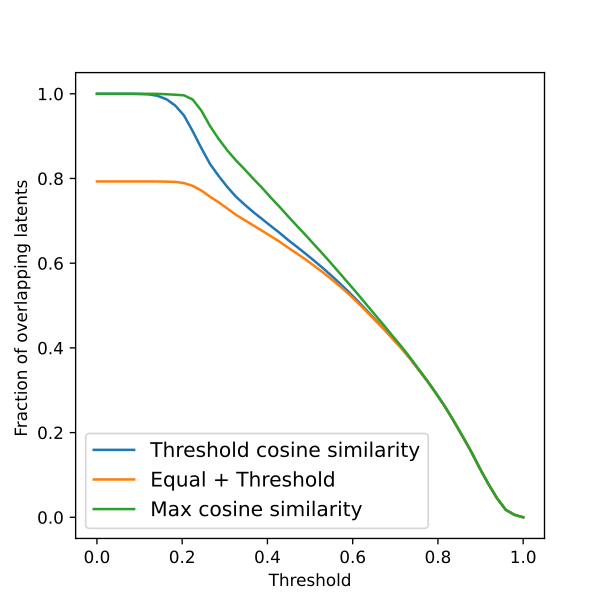}
    \caption{The average alignment of points with equal decoder and encoder indices is 0.72 and of the ones that have different indices is 0.33. On the right, we plot the fraction of latents that are considered shared between 2 SAEs as we control a threshold. We decide to use a threshold of 0.7 on both the encoder and decoder alignment to decide if a latent is shared between two SAEs.}
    \label{fig:aligned_label}
\end{figure}

\begin{figure}[h]
    \centering
    \includegraphics[width=0.6\linewidth]{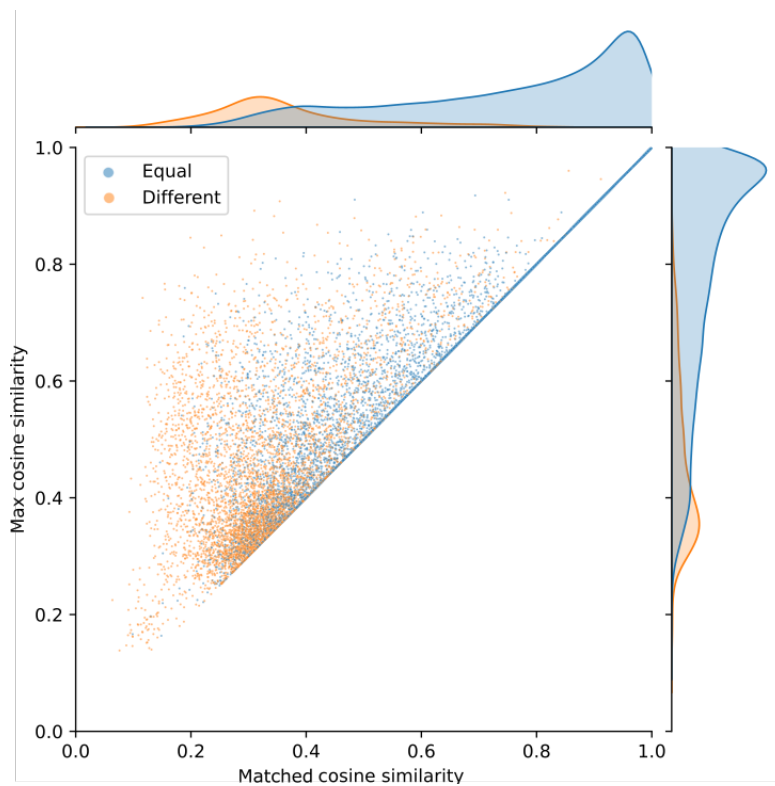}
    \caption{\textbf{Cosine similarity of latents when paired with the Hungarian algorithm vs when using max cosine similarity}. The majority of latents that have the same counterpart feature in both the encoder and decoder matchings using the Hungarian algorithm have a similar alignment as if they had been aligned with maximum cosine similarity. The latents which have a higher cosine similarity pair when using max cosine similarity are paired with a latent that already had a pair. 
    }
    \label{fig:matched_max_cosine_scatter}
\end{figure}

\begin{figure}[h]
    \centering
    \includegraphics[width=1\linewidth]{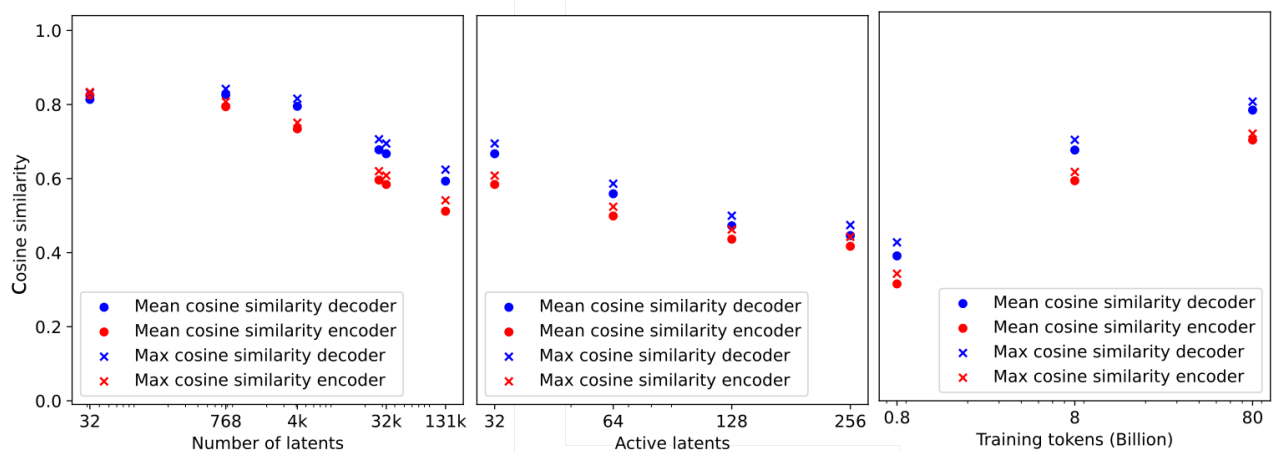}
    \caption{\textbf{Dependence of mean matched and mean max cosine sim of a Pythia-160M SAE on different hyperparameters}. On the left we see that the average cosine similarity of latents decreases with the increase of the total number of latents. Middle shows that increasing the number of active latents also decreases the average cosine similarity.  On the right, training time increases the average cosine similarity of different SAE seeds. We observe that the mean matched and max cosine similarity have very similar trends, with max cosine similarity being just slightly higher. On panels all panels a 32768 latent SAE was trained on the output MLP of Pythia 160M, for 8B tokens, except when the panel changes one of these conditions. 
}
    \label{fig:matched_max_cosine}
\end{figure}
%%%%%%%%%%%%%%%%%%%%%%%%%%%%%%%%%%%%%%%%%%%%%%%%%%%%%%%%%%%%%%%%%%%%%%%%%%%%%%%
%%%%%%%%%%%%%%%%%%%%%%%%%%%%%%%%%%%%%%%%%%%%%%%%%%%%%%%%%%%%%%%%%%%%%%%%%%%%%%%

\end{document}